\documentclass[manuscript,screen,preprint]{acmart}

\usepackage{tabu}
\usepackage{tabularx}
\usepackage{amsmath}
\usepackage{bbm}
\usepackage{graphicx}
\usepackage{caption}
\usepackage{subcaption}
\usepackage{hyperref}
\usepackage{url}
\usepackage{cases}
\usepackage[english]{babel}
\usepackage{blindtext}
\usepackage{bm}
\usepackage{multirow}
\usepackage{makecell}
\usepackage{balance}

\usepackage{algorithm}
\usepackage{algpseudocode}
\usepackage{xcolor}
\usepackage{rotating}

\renewcommand\footnotetextcopyrightpermission[1]{} 
    
\begin{document}


\title{Tuning Language Models for Robust Prediction of Diverse User Behaviors}


\author{Fanjin Meng}
\authornote{Both authors contributed equally to this research.}
\affiliation{%
  \institution{Department of Electronic Engineering, BNRist, Tsinghua University}
  \country{Beijing, China}}
\email{mengfj23@mails.tsinghua.edu.cn}

\author{Jingtao~Ding$^{*}$}

\affiliation{%
  \institution{Department of Electronic Engineering, BNRist, Tsinghua University}
  \country{Beijing, China}}
\email{dingjt15@tsinghua.org.cn}

\author{Jiahui Gong}
\affiliation{%
  \institution{Department of Electronic Engineering, BNRist, Tsinghua University}
  \country{Beijing, China}}
\email{gjh22@mails.tsinghua.edu.cn}

\author{Chen Yang}
\affiliation{%
  \institution{Honor Device Co., Ltd	}
  \country{Beijing, China}}
\email{yangchen6@honor.com}

\author{Hong Chen}
\affiliation{%
  \institution{Honor Device Co., Ltd	}
  \country{Beijing, China}}
\email{chenhong3@honor.com}

\author{Zuojian Wang}
\affiliation{%
  \institution{Honor Device Co., Ltd	}
  \country{Beijing, China}}
\email{wangzuojian@honor.com}

\author{Haisheng Lu}
\affiliation{%
  \institution{Honor Device Co., Ltd	}
  \country{Beijing, China}}
\email{luhaisheng@honor.com}

\author{Yong Li}
\affiliation{%
  \institution{Department of Electronic Engineering, BNRist, Tsinghua University}
  \country{Beijing, China}}
\email{liyong07@tsinghua.edu.cn}

\renewcommand{\shortauthors}{Fanjin Meng et al.}

\begin{abstract}
Predicting user behavior is essential for intelligent assistant services, yet deep learning models often struggle to capture long-tailed behaviors. Large language models (LLMs), with their pretraining on vast corpora containing rich behavioral knowledge, offer promise. However, existing fine-tuning approaches tend to overfit to frequent ``anchor'' behaviors, reducing their ability to predict less common ``tail'' behaviors.
In this paper, we introduce BehaviorLM, a progressive fine-tuning approach that addresses this issue. In the first stage, LLMs are fine-tuned on anchor behaviors while preserving general behavioral knowledge. In the second stage, fine-tuning uses a balanced subset of all behaviors based on sample difficulty to improve tail behavior predictions without sacrificing anchor performance.
Experimental results on two real-world datasets demonstrate that BehaviorLM robustly predicts both anchor and tail behaviors and effectively leverages LLM behavioral knowledge to master tail behavior prediction with few-shot examples. 
Code and data are available at \color{blue}\url{https://anonymous.4open.science/r/1015-new-E5F0}
\end{abstract}


\maketitle
\begingroup\renewcommand\thefootnote{\textsection}
\footnotetext{Corresponding Author.}

\section{Introduction}

Most people lead routine lives, driven by behavioral habits, but also exhibit short-term bursts of activity influenced by specific contexts. The ability to predict users' next behavior is crucial for modern intelligent assistant services, which span a wide range of applications, from web platforms to smart devices~\cite{chung2018intelligent,tulshan2019survey,savcisens2023using,rychalska2023synerise}.

With the accumulation of empirical data on user behaviors, deep learning-based approaches have replaced traditional rule-based methods~\cite{zhang2018mobility,zhang2019deep,li2022smartphone} and become the mainstream solution in this field. Among them, transformer-based models~\cite{kang2018self,Pu19transformer,sun2019bert4rec,savcisens2023using} excel at capturing transition patterns between sequential behaviors. However, sequential behavior modeling typically demands extensive training data, which incurs significant costs. In this context, large language models (LLMs)~\cite{GPT3,zhao2023survey} have emerged as an ideal choice for behavior prediction, as these pretrained models inherently encode knowledge about human behavior from their vast training corpora.  They show promise in interpreting the underlying intent behind observed actions and generating corresponding predictions~\cite{wu2024survey,zhao2024recommender}.

To adapt LLMs for behavior prediction, current approaches typically convert behavioral sequence data into textual format and fine-tune LLMs to predict user behavior tokens~\cite{li2023text,bao2023tallrec,liao2024llara,kimlarge}. Among them, an early attempt~\cite{bao2023tallrec} combines general instruction tuning on conversation data with lightweight fine-tuning on behavior data. Building on this, subsequent research~\cite{liao2024llara} enhances LLMs' behavioral understanding by incorporating traditional models' encoded embeddings into text instructions. While expressing user behaviors as text appears intuitive~\cite{li2023text}, the long-tailed distribution of behaviors presents a significant challenge~\cite{liu2024llm}. Specifically, a user's daily life is predominantly characterized by a small subset of frequent behaviors serving as ``anchors'', while other behaviors occur far less commonly.
Recent work~\cite{liu2024llm} proposes combining LLMs' capacity to characterize item profiles, particularly for long-tailed items, with traditional sequential recommendation modules. However, capturing this long-tailed distribution remains a persistent challenge for LLMs. 
Our empirical observations reveal that an 8B-parameter LLM, after behavior data fine-tuning, readily outperforms the more powerful but untuned GPT-4 on anchor behaviors, yet falls short on tail behaviors. This disparity suggests that fine-tuning alone does not uniformly enhance prediction capabilities across diverse behaviors.

In this paper, we empirically investigate LLMs' uneven performance across diverse behaviors after fine-tuning and, surprisingly, find that an LLM fine-tuned exclusively on anchor behaviors can effectively predict tail behaviors in a zero-shot manner. Inspired by this novel finding, we propose a progressive fine-tuning approach to better capture the long-tailed distribution of user behavior. In the first stage, to preserve general behavioral knowledge and prevent bias toward overrepresented behaviors, we fine-tune the LLM not only with anchor behaviors but also by integrating a general conversational corpus in a multi-task manner. This allows the LLM to specialize in predicting anchor behaviors while retaining the ability to generalize to unseen tail behaviors. In the second stage, we further fine-tune the LLM across all behavior types by selecting a balanced subset of data based on sample difficulty. This approach significantly enhances prediction accuracy for tail behaviors without compromising performance on anchor behaviors. This two-stage fine-tuning framework, which we call BehaviorLM, effectively leverages the LLM's general behavioral knowledge for sample-efficient tuning, enabling robust predictions across both anchor and tail behaviors.

To summarize, our main contributions are as follows.
\begin{itemize}
    \item We introduce a novel approach for adapting LLMs to the human behavioral domain, addressing the longstanding challenge of capturing long-tailed user preferences.
    
    \item We propose a progressive LLM fine-tuning strategy, inspired by empirical findings, that effectively leverages the general behavioral knowledge stored in LLMs for behavior prediction.

    \item Experimental results on two real-world datasets demonstrate the superiority of BehaviorLM over state-of-the-art baselines in behavior prediction. Notably, BehaviorLM achieves a {30.8\%/22.5\%} improvement in prediction accuracy for tail behaviors, underscoring its ability to capture long-tailed preferences. In-depth analysis reveals that the behavioral knowledge stored in the LLM provides over a 100$\times$ improvement in sample efficiency compared to traditional transformer-based approaches, while also enabling few-shot~($\sim$ 20) predictive capability for tail behaviors.  Ablation studies further validate the rationale behind the design of our method.

\end{itemize}

\section{Motivation}

\begin{figure*}[t]
    \centering
    \includegraphics[width = 1.0 \linewidth]{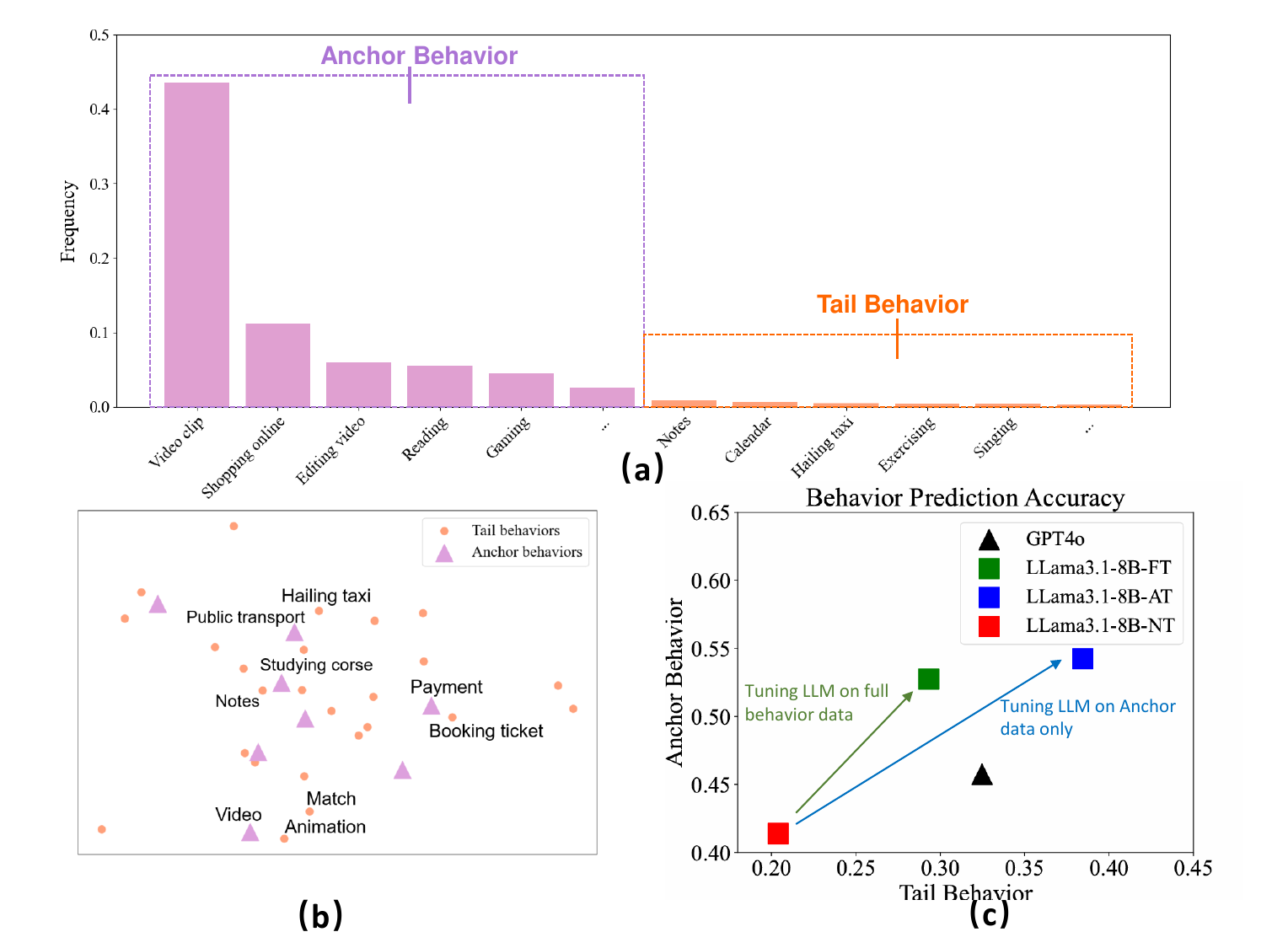}
    \vspace{-2mm}
    \caption{(a) Empirical distribution of user behaviors in the Behavior dataset: "Anchor Behaviors" occur more than 1\% of the time, while "Tail Behaviors" represent the rest. (b) Semantic embedding visualization of anchor and tail behaviors in the LLM. (c) Prediction accuracy comparison across LLM tuning methods and GPT4o for anchor and tail behaviors, with "NT" indicating no tuning.}
    \vspace{-2mm}
    \label{fig:motivation}
    \end{figure*}

We first give a formal definition of our research problem and then present our novel observations that motivate the methodology design of BehaviorLM.

\subsection{Problem Formulation} 

\textbf{Next behavior prediction} 
refers to the task of predicting a user's next behavior, $y \in \mathcal{B}$, given a chronologically ordered sequence $x = \{ e_1, e_2, \dots, e_L \}$ of their most recent $L$ historical behavioral events. Each event $e = \left( l, t, b \right)$ indicates that a specific behavioral event $b \in \mathcal{B}$ occurred at location $l$ and time $t$. 
The behavioral event $b$ specifically refers to daily activities—such as exercise or gaming—rather than fine-grained actions like picking up a cup.
The location $l$ 
{indicates a semantic indicator such as "home" or "workplace."} The time $t$ includes both date and hour information.

Given a collected dataset $\mathcal{D} = \left\{ (x_i, y_i) \right\}_{i=1,\dots,N}$, our goal is to train a prediction model $M_{\Phi}$ capable of predicting the next behavioral event, i.e., $y = M_{\Phi}(x).$

To facilitate understanding, we summarize the key mathematical notations used in this paper in Table~\ref{tab:notations}.

\begin{table}[h]
    \centering
    \caption{The description of notations.}
    \label{tab:notations}
    \begin{tabular}{c|l}
        \toprule
        Notations & Description \\
        \midrule
        $\mathcal{B}$ & The set of all possible user behaviors (candidate set) \\
        $x$ & The historical behavior sequence of a user \\
        $y$ & The ground truth next behavior to be predicted \\
        $e$ & A specific behavioral event tuple (location, time, behavior) \\
        $L$ & The length of the historical behavior sequence \\
        $\mathcal{D}$ & The collected raw user behavior dataset \\
        $\mathcal{D}_{\text{ins}}$ & The instruction tuning dataset converted from $\mathcal{D}$ \\
        $\mathcal{D}_{\text{ins}}^a$ & The subset of anchor behaviors (frequent behaviors) \\
        $\mathcal{D}_{\text{ins}}^t$ & The subset of tail behaviors (infrequent behaviors) \\
        $\mathcal{C}_{\text{ins}}$ & The auxiliary conversation dataset for multi-task learning \\
        $\epsilon$ & Ratio controlling the size of auxiliary task data \\
        $\Phi$ & The learnable parameters of the LLM \\
        $\pi_{\text{ref}}$ & The reference model obtained from the A-Tuning stage \\
        $\pi_{\theta}$ & The policy model being optimized in the B-Tuning stage \\
        $p(y \mid x)$ & Prediction confidence of the model for behavior $y$ \\
        $d_{\text{confusion}}$ & The confusion-based penalty score \\
        $D(x)$ & The continuous difficulty coefficient for a sample \\
        $\lambda$ & Hyper-parameter balancing uncertainty and confusion penalty \\
        $\beta$ & Hyper-parameter controlling deviation in DPO loss \\
        $\mathcal{L}_{\text{DPO}}$ & The Direct Preference Optimization (DPO) loss function \\
        \bottomrule
    \end{tabular}
\end{table}

\subsection{Investigating LLM Fine-tuning Performance on Diverse User Behavior Dataset}

In this subsection, we empirically investigate the behavior prediction performance of current fine-tuning approaches for LLMs. We utilize a real-world dataset recording 37 types of daily user behaviors on a smartphone (detailed in Section 4). As shown in Figure~\ref{fig:motivation}(a), there is a significant class-imbalance issue in terms of occurrence frequency among different behavior types. Without loss of generality, we divided these behaviors into two categories: anchor behaviors, with an occurrence frequency greater than 1\%, and tail behaviors for the rest. Although there are only 16 types of anchor behaviors, they account for over 97\% of the total data, with an average ratio between anchor and tail behaviors of approximately 42.44 (0.97/16 vs. 0.03/21), and the highest ratio exceeding 2,500 (43.6\% vs. 0.0163\%). 
 
To further analyze the semantic meaning and similarity of these diverse behaviors from the perspective of the LLM, we use a pretrained LLM, Llama-8B v3.1~\cite{dubey2024llama}, to generate embedding vectors for the behaviors and visualize them in two-dimensional space (reduced from 4096 dimensions) using PCA~\cite{abdi2010principal}. As shown in Figure~\ref{fig:motivation}(b), anchor behaviors act as semantic anchors in the latent space, scattered across it, with several tail behaviors clustering around these anchor points. For example, watching videos, especially short videos, has become a highly frequent behavior in users' daily lives, while other nearby behaviors in the figure, such as watching sports matches or animation, occur less frequently and are favored by fewer individuals. Similarly, public transportation is a more common commuting behavior than hailing a taxi for most people.

To investigate the behavior prediction performance of LLMs, we use Llama-8B v3.1 as the backbone model and fine-tune it on the behavior dataset described earlier. For evaluation, we select an equal number of samples for each behavior type and report the average prediction accuracy for anchor and tail behaviors, respectively, as shown in Figure~\ref{fig:motivation}(c). For comparison, we first evaluate the prediction accuracy of GPT-4 (version gpt-4o-2024-08-06), which yields accuracy values of 0.45 for anchor behaviors and 0.33 for tail behaviors. 
{In its base form, LLaMA3.1-8B significantly underperforms GPT-4 in predicting both anchor and long-tail behaviors. After applying established fine-tuning practices~\cite{bao2023tallrec,liao2024llara}, the model shows marked improvement on anchor behaviors, surpassing GPT-4's performance. However, for long-tail behaviors, it continues to lag behind GPT-4's capabilities.
GPT-4's superior comprehensive performance across all behavior types, achieved without behavior-specific fine-tuning, suggests that current fine-tuning approaches may disproportionately optimize for anchor behaviors at the expense of tail behaviors. We attribute this performance disparity to the severe class imbalance illustrated in Figure~\ref{fig:motivation}(a).}

Motivated by these observations, we investigated whether eliminating class imbalance during fine-tuning could better leverage the LLM's inherent knowledge of user behaviors.
We fine-tuned the LLM exclusively on anchor behavior data, excluding all tail behavior samples. As illustrated in the figure, this approach maintains strong performance on anchor behaviors. Remarkably, despite having no exposure to tail behaviors during fine-tuning, the model demonstrates robust zero-shot predictive capabilities for these behaviors, achieving an accuracy of 0.39—substantially higher than the 0.29 obtained through traditional fine-tuning. This finding suggests that training on anchor behaviors alone enables the LLM to develop a fundamental understanding of user behavior patterns and leverage its general knowledge to generalize to unseen, semantically related behaviors. This insight forms the foundation of our progressive fine-tuning approach for BehaviorLM.

\section{Method}
\subsection{The BehaviorLM Framework}
\subsubsection{Behavior prediction as language modeling}
To transform the next behavior prediction task into a language modeling task, we first design the specific prompt and then adopt the instruction fine-tuning technique for LLMs.

\textbf{Prompt design.}  
We adopt a text-only prompt design that represents the behavioral history $x$ and the next behavior $y$ (if given) through textual metadata embedded within the prompts, allowing user behavior data to be transformed into training data suitable for LLM instruction tuning. The prompt consists of the following five parts:

\begin{enumerate}
    \item \textbf{Task definition}: A brief description of the next behavior prediction task, e.g., predict the next behavior the user will do from the candidate.
    \item \textbf{Role-playing instruction}: This defines the expected role and task for the LLM to assume and complete~\cite{shanahan2023role}, e.g., you are a smart user's mobile phone assistant, which can infer the user's mobile phone behavior preferences based on the historical behavior history of the user.
    \item \textbf{Historical behavior sequence (Input)}: Each element in $x$ is replaced with a textual description, e.g.,(1,16,home,Exercise), ...,(1,20,home,Online shopping),(1,20,home,Video).
    \item \textbf{Candidate set (Input)}: A collection of all possible behaviors the LLM can predict, constraining the range of behaviors to choose from, e.g., Weather Check, Music,..., Cycling, Reading.
    \item \textbf{Next behavior (Output)}: The ground truth label $y$ for the LLM to predict, e.g., Gaming.
\end{enumerate}

\textbf{Instruction Fine-Tuning LLMs.}  
To quickly adapt an LLM to behavior prediction tasks, we adopt instruction fine-tuning techniques~\cite{ouyang2022training}. Based on the above prompt design, we transform the original user behavior dataset $\mathcal{D}$ into a text-based instruction dataset $\mathcal{D}_\text{ins} = \left\{ (x^p_i, y^p_i) \right\}_{i=1,\dots, N}$, where $x^p_i$ and $y^p_i$ represent the corresponding natural language form of the input and output.
The LLM is then optimized using the following next-token prediction loss:
\begin{equation}
   \max_{\Phi} \sum_{(x,y) \in \mathcal{D}_\text{ins}} \sum_{s=1}^{|y|} \log \left( P_{\Phi}(y_s \mid x, y_{<s}) \right), 
\end{equation}
where $\Phi$ represents the learnable parameters of the LLM, $y_s$ refers to the $s$-th token of $y$, and $y_{<s}$ indicates all tokens preceding $y_s$.
Moreover, to efficiently fine-tune the LLM with its vast number of parameters, we adopt a well-known parameter-efficient fine-tuning technique, i.e., LoRA~\cite{hu2021lora}. LoRA freezes the original parameter weights, decomposing them into the original part $\Phi_0$ (which remains frozen) and an additional low-rank matrix $\Delta \Phi$, which can be updated much more efficiently while maintaining the performance of the LLM.
By framing the task of predicting the next user behavior as predicting the next language token, we can more effectively leverage the LLM's knowledge from its pretraining stage.

\subsubsection{Progressive fine-tuning with behavior data}
As shown in Figure~\ref{fig:motivation}(a), user behavior data exhibits a significant \textit{class imbalance issue}, where many types of behaviors occur infrequently in daily life, resulting in a long-tailed distribution. When transforming the task of predicting a user's next behavior into generating the corresponding language token, the fine-tuned LLM still suffers from this class imbalance, leading to poor prediction performance for long-tailed behaviors, as demonstrated by ``Llama3.1-8B-FT'' in Figure~\ref{fig:motivation}(b).
Motivated by the discovery that an LLM fine-tuned on a subset of frequently occurring behaviors (anchor behaviors) can surprisingly serve as an effective zero-shot predictor for other, less frequent behaviors (tail behaviors), {performing even better than fine-tuning on the full set of behavioral data}, we design a progressive fine-tuning strategy to enhance BehaviorLM's predictive capability across diverse user behaviors.

The proposed fine-tuning strategy consists of two progressive stages (Figure~\ref{fig:framework}). In the first stage, the LLM is fine-tuned using anchor behavior data to specialize in user behavior prediction while retaining its inherent rich knowledge of long-tailed behaviors. In the second stage, the model is further fine-tuned on a class-balanced dataset covering all behaviors in a few-shot manner, enhancing its predictive capability for tail behaviors. The first stage helps the LLM become a specialist in predicting anchor behaviors, while the second stage transforms it into a generalist capable of predicting both anchor and tail behaviors.

\begin{figure*}[t]
    \centering
    \includegraphics[width = 1.0\linewidth]{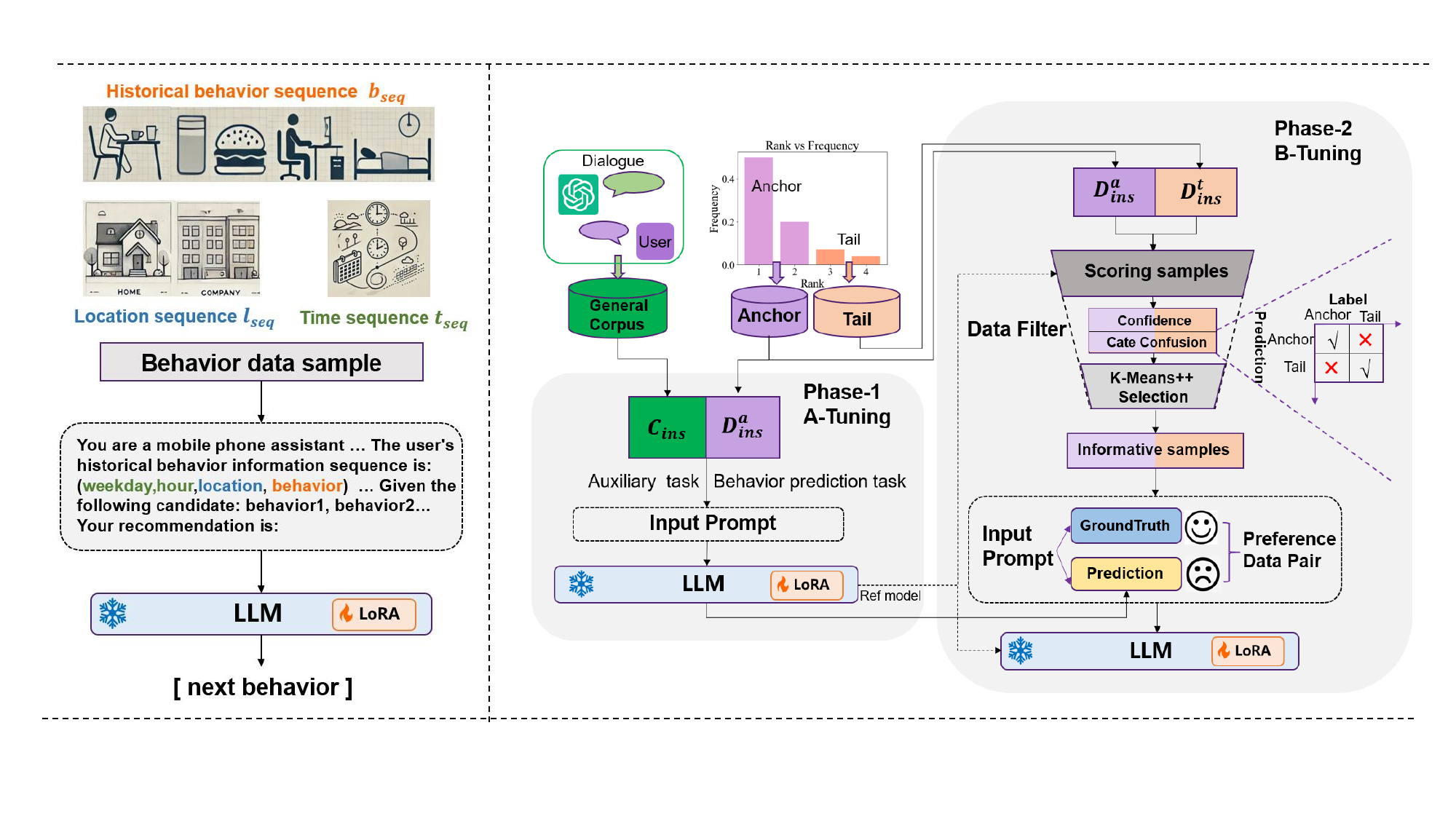}
    \vspace{-4em}
    \caption{ The BehaviorLM framework, with a progressive fine-tuning approach}
    \label{fig:framework}
    \end{figure*}

\subsection{Multitask-Enhanced Anchor Behavior Tuning (A-Tuning)}

We divide the instruction fine-tuning data $\mathcal{D}_\text{ins}$ into two parts: $\mathcal{D}^a_\text{ins}$, which contains labels belonging to anchor behaviors, and $\mathcal{D}^t_\text{ins}$, which contains the rest. In this stage, we fine-tune the LLM using only $\mathcal{D}^a_\text{ins}$.
As shown in Figure~\ref{fig:motivation}, anchor behaviors represent the core patterns of a user's daily life. Therefore, after fine-tuning with anchor behaviors, the LLM gains a preliminary understanding of the underlying patterns in user behavior, enabling it to accurately predict the next anchor behavior. Moreover, since the LLM already captures general behavioral knowledge from its pretraining corpus, fine-tuning with anchor behaviors allows it to generalize more easily to other unseen but semantically similar behaviors.
This approach, compared to fine-tuning on the entire behavior dataset $\mathcal{D}_\text{ins}$, helps prevent the LLM from being biased towards a few behavior types that are overrepresented in the data. 

To further enhance the LLM's generalization from anchor to tail behaviors, we propose a multi-task fine-tuning approach. Beyond learning to predict next behaviors in $\mathcal{D}^a_\text{ins}$, we maintain the model's general task-solving capabilities—a strategy our experiments later confirmed to be advantageous for behavior prediction tasks. To achieve this, we simultaneously fine-tune the LLM on another dataset, $\mathcal{C}_\text{ins}$, derived from daily conversations between users and ChatGPT~\cite{zheng2023judging}. This effectively integrates an auxiliary task of conversation generation alongside the primary behavior prediction task. We control the impact of the auxiliary task by adjusting the size of $\mathcal{C}_\text{ins}$, i.e., using a ratio $\varepsilon$ relative to $\mathcal{D}^a_\text{ins}$.
In practice, we filter out excessively long conversations from $\mathcal{C}_\text{ins}$ to ensure that the prompts from both tasks are comparable in length.

\subsection{Difficulty-based Data Selection for Balanced Behavior Tuning (B-Tuning)}

In the second stage, we reintroduce the tail behaviors $\mathcal{D}^t_\text{ins}$ and combine them with $\mathcal{D}^a_\text{ins}$ to create a class-balanced fine-tuning dataset. Since the LLM fine-tuned during the A-Tuning stage already serves as a good zero-shot predictor for tail behaviors, we believe that a small amount of fine-tuning data covering all behavior types should suffice to build a robust user behavior predictor. However, to achieve this, the quality and informativeness of the selected samples play a crucial role. Specifically, we designed a sample difficulty scoring strategy, searching for difficult samples from the following two dimensions for effectively fine-tuning the LLM in a few-shot way.

\textbf{Confidence-based difficulty.}
One simple way to measure sample difficulty is by scoring samples with an intermediate model, and those with the wrong predicted labels are more difficult than those correctly predicted~\cite{bengio2009curriculum}. In our implementation, we directly use the fine-tuned LLM from the A-Tuning stage as the sample scorer and compute the prediction confidence $p(y \mid x)$ for the ground truth behavior $y$. A lower confidence score indicates that the model struggles to model the user's intention correctly. This forms the first component of our difficulty score: $1 - p(y \mid x)$.

\textbf{Confusion-based penalty.}
Following the idea of contrastive learning~\cite{liu2021self}, we further select difficult samples by considering the distinguishability between their predicted labels and groundtruth labels. We aim to choose those mispredicted samples with a lower distinguishability score $d_{confusion}$(the second component of our difficulty score), which is calculated as follows, 
\begin{equation}
    d_{confusion} = 
    \begin{cases}
        0, & \text{if } \hat{y} \neq y \text{ and } \mathrm{Cate}(\hat{y}) = \mathrm{Cate}(y) \\
        1, & \text{otherwise}
    \end{cases}
\end{equation}

where $\hat{y}$ denotes the predicted behavior, $y$ denotes the ground truth label, and $\text{Cate}(y)$ denotes behavior category of $y$, i.e., anchor or tail.
This follows the idea that anchor behaviors (tail behaviors) are much more difficult to discriminate from themselves than their counterparts.

Finally, we combine the above two dimensions to define the final continuous difficulty coefficient $D(x)$ for each sample:
\begin{equation}
    D(x) = \lambda \cdot (1 - p(y \mid x)) + (1 - \lambda) \cdot d_{confusion} 
\end{equation}
where $\lambda \in [0, 1]$ balances the confidence uncertainty and the semantic confusion penalty, its default value in the experiment was set to 0.5.

\textbf{Difficulty-Weighted K-Means++ Selection.}
While the continuous difficulty coefficient $D(x)$ provides a nuanced measure of hardness, relying solely on difficulty magnitude for selection creates a new challenge: the selected hard samples might be redundant (e.g., clustering in a specific failure mode). To address this and construct a dataset that is both \textit{challenging} and \textit{representative}, we map all samples into a semantic space using the LLM's embeddings. For each behavior category, we apply the Difficulty-Weighted K-Means++ algorithm to select a fixed number $F$ of samples to create the class-balanced fine-tuning dataset\footnote{Since $F$ is usually small, we do not have to score each sample with the LLM to obtain suitable samples, which can be time-costly.}. Crucially, we use the normalized difficulty score $\tilde{D}(x) = D(x) / \max_{x'} D(x')$ as the sampling probability weight during cluster initialization. This mechanism ensures that the sampled data covers the diverse semantic landscape of user behaviors while preferentially targeting those instances where the model exhibits high uncertainty or confusion.

\textbf{Preference Optimization with DPO.}
Unlike previous A-Tuning Stage that use Supervised Fine-Tuning (SFT), we adopt DPO to further align the model to make fuller use of the selected high-value hard samples. We construct preference pairs $(x, y_w, y_l)$ where the ground truth is the chosen response $y_w = y$, and the incorrect prediction from the A-Tuning model serves as the rejected response $y_l = \hat{y}$. The objective is to minimize the DPO loss:
\begin{equation}
    \mathcal{L}_{\text{DPO}} = - \mathbb{E}_{(x, y_w, y_l) \sim \mathcal{D}_{\text{dpo}}} \left[ \log \sigma \left( \beta \log \frac{\pi_\theta(y_w|x)}{\pi_{\text{ref}}(y_w|x)} - \beta \log \frac{\pi_\theta(y_l|x)}{\pi_{\text{ref}}(y_l|x)} \right) \right]
\end{equation}
where $\pi_{\text{ref}}$ is the frozen model from A-Tuning, $\pi_\theta$ is the policy model being optimized, and $\beta$ controls the deviation from the reference model.

By adopting the data selection strategy that balances difficulty and diversity, combined with the DPO training paradigm, we can prompt the LLM to capture more refined behavior knowledge in a sample-efficient manner and significantly enhance its prediction accuracy across tail behaviors without compromising those of anchor behaviors.

We also summarize the designed progressive fine-tuning strategy for BehaviorLM in Algorithm~\ref{alg}.

\begin{algorithm}[t]
\caption{BehaviorLM: Progressive Fine-tuning Strategy}
\label{alg}
\begin{algorithmic}[1]
    \Require 
    Anchor Dataset $\mathcal{D}^a_\text{ins}$, Tail Dataset $\mathcal{D}^t_\text{ins}$, Auxiliary Conversation Dataset $\mathcal{C}_\text{ins}$; \\
    Auxiliary ratio $\varepsilon$, Balance weight $\lambda$, Samples per class $F$, DPO beta $\beta$.
    \Ensure Optimized LLM $\pi_\theta$.

    \Statex \textbf{\textit{Stage 1: A-Tuning (SFT)}}
    \State Sample auxiliary conversation data $\mathcal{C}'_\text{ins} \subset \mathcal{C}_\text{ins}$ where $|\mathcal{C}'_\text{ins}| = \varepsilon \cdot |\mathcal{D}^a_\text{ins}|$
    \State Construct multi-task dataset $\mathcal{D}_\text{SFT} \leftarrow \mathcal{D}^a_\text{ins} \cup \mathcal{C}'_\text{ins}$
    \State $\pi_{\text{ref}} \leftarrow \text{LoRA-SFT}(\mathcal{D}_\text{SFT})$ \Comment{Obtain reference model}

    \Statex \textbf{\textit{Stage 2: B-Tuning (DPO)}}
    \State Initialize DPO dataset $\mathcal{D}_\text{DPO} \leftarrow \emptyset$
    \State $\mathcal{D}_\text{pool} \leftarrow \mathcal{D}^a_\text{ins} \cup \mathcal{D}^t_\text{ins}$
    
    \State \textbf{Step 2.1: Difficulty Scoring \& Semantic Embedding}
    \For{sample $x \in \mathcal{D}_\text{pool}$}
        \State Get semantic embedding $E(x)$
        \State Predict $\hat{y}$ and confidence $p(y|x)$ using $\pi_{\text{ref}}$
        \State Compute confusion penalty $d_{confusion}$ (Eq. 2)
        \State Calculate difficulty score $D(x)$ (Eq. 3)
    \EndFor

    \State \textbf{Step 2.2: Difficulty-Weighted Sampling}
    \For{each behavior category $c$}
        \State Retrieve samples $\mathcal{X}_c = \{x \in \mathcal{D}_\text{pool} \mid \text{Cate}(x) = c\}$
        \State Calculate normalized weights $\tilde{D}(x) \propto D(x)$ for $x \in \mathcal{X}_c$
        \State $\mathcal{S}_c \leftarrow \text{KMeans++}(\text{data}=\mathcal{X}_c, \text{weights}=\tilde{D}, k=F)$ \Comment{Select representative hard samples}
        
        \State \textbf{Step 2.3: Preference Pair Construction}
        \For{each selected sample $x \in \mathcal{S}_c$}
            \State Set chosen response $y_w \leftarrow y$ (Ground Truth)
            \State Set rejected response $y_l \leftarrow \hat{y}$ (Prediction from $\pi_{\text{ref}}$)
            \State $\mathcal{D}_\text{DPO} \leftarrow \mathcal{D}_\text{DPO} \cup \{(x, y_w, y_l)\}$
        \EndFor
    \EndFor

    \State \textbf{Step 2.4: Direct Preference Optimization}
    \State $\pi_\theta \leftarrow \text{LoRA-DPO}(\pi_{\text{ref}}, \mathcal{D}_\text{DPO}, \beta)$ \Comment{Minimize $\mathcal{L}_{\text{DPO}}$ (Eq. 4)}
    
    \State \Return $\pi_\theta$
\end{algorithmic}
\end{algorithm}

\section{Experiment}
\subsection{Experiment Settings}
\subsubsection{Datasets.}
We evaluated our model on two real-world user behavior datasets:

\textbf{Honor behavior dataset}: This large-scale dataset is derived from mobile device \footnotemark usage logs. When users interact with their mobile phones, various types of logs are generated, desensitized, and reported with user consent. We select 37 daily behaviors that are reliably extracted from raw logs and also cover broad life scenarios, including activities related to learning, work, entertainment, leisure, etc. The dataset spans from March 1, 2024, to April 29, 2024, and consists of over 50 million behavior events from 24,133 anonymous users. We preprocess the dataset and construct samples in the format of ``(last-20 events, next event)''. Since our target is fine-tuning the LLM instead of training from scratch, we only randomly select a subset~(200,000) of data for experiments.

\textbf{App usage dataset~\cite{li2022smartphone}}: This dataset is an open-source resource that captures 1753 user interactions across various apps within one week. Given the large number of apps and the overlap in functionality among many of them, we processed the dataset further by merging apps with similar purposes. For example, Douyin and Kuaishou were grouped into a "watching videos" category. After similar preprocessing as Honor dataset, this dataset consists of 71 behavior events belonging to 24 categories.

The characteristics of both datasets are presented in Table~\ref{datasets-infor}, and the detailed information, including behavior types for the Honor behavior dataset, is provided in Appendix~\ref{app:dataset}.

\footnotetext{https://www.honor.com/global/}

\begin{table}
  \centering
  \caption{Statistics of the datasets}
  \label{datasets-infor}
  \resizebox{0.75\textwidth}{!}{
  \begin{tabular}{cccc}
    \toprule
    Dataset      & \# User   &   \# Behavior Type &\# Sample  \\
    \midrule
    Honor Behavior Dataset   &24,133 &  37 &   200,000    \\
    App Usage Dataset   &1,753 & 24 &  50,000       \\
    \bottomrule
  \end{tabular}
  }
  \label{tab:three-line}
\end{table}

\subsubsection{Evaluation Protocols and Metrics}
To comprehensively evaluate our model, we consider two different evaluation protocols.
First, we follow the common practice of \textit{next behavior prediction}~\cite{bao2023tallrec,liao2024llara} to evaluate the overall prediction performance of the model. {First, we divided the users in an 8:1:1 ratio and applied consistent data processing methods to generate the training set, validation set, and test set samples.}
Second, since we aim to evaluate whether the model can robustly predict across both anchor and tail behaviors, we also follow common practice of \textit{long-tailed learning}~\cite{liu2019large,shi2024long} to construct an additional test set, where each behavior category contains equally 500 samples, ensuring that rarely-occurred tail behaviors are sufficiently evaluated compared with the first \textit{next behavior prediction} protocol that reflects the real-world distribution of behavior categories. Note that this additional test set is extracted from the original Hornor dataset and App dataset, which were not used in the first protocol. For each evaluation protocol, we repeat experiments five times and report mean values.

Overall, we adopt six commonly used metrics. The \textit{next behavior prediction} evaluation considers weighted precision ($Prec_w$) and weighted recall ($Rec_w$), accounting for the overall performance by considering the proportion of behavior categories. {Note that the recall calculation ($Rec_w$) used here is equivalent to the widely adopted recommendation metric Hit-Rate$@1$.} The \textit{long-tailed learning} evaluation considers four specific accuracy metrics according to the occurrence frequency of different behaviors. These are category-average accuracy for all behavior types (denoted as $Overall$), head-category accuracy for whose frequency larger than 5.0\% (denoted as $Head$), medium-category accuracy for whose frequency larger than 1.0\% and lower than 5.0\% (denoted as $Medium$), and tail-category accuracy for the rest behavior types (denoted as $Tail$), { following the settings of ~\cite{liu2019large,shi2024long}}. By utilizing these metrics, we focus on whether the model can make robust predictions across diverse behaviors and better assess its predictive performance on imbalanced behavioral datasets. The specific calculation formula for the above six metrics is listed in Appendix~\ref{app:metric}.

\subsubsection{Baselines}
We selected the following seven representative algorithms to compare with our proposed algorithm, covering traditional methods (SASRec~\cite{kang2018self} and Bert4Rec~\cite{sun2019bert4rec}), LLM-enhanced methods ({PITuning~\cite{gong2024population}, LLM-ESR~\cite{liu2024llm} and AlphaFuse~\cite{hu2025alphafuse}), LLM-based methods (GPT4o~\cite{achiam2023gpt}, A-LLMRec~\cite{kimlarge}, TALLRec~\cite{bao2023tallrec}, LLaRa~\cite{liao2024llara}) and CoLLM~\cite{zhang2025collm}}. 
The details of baselines are provided in Appendix~\ref{app:baselines}.

\subsubsection{Implementation Details.}
We selected LLama3.1-8B~\cite{dubey2024llama} as the backbone for our experiments. To ensure flexibility in model testing, we designed three distinct instruction formats, which are randomly sampled during both training and testing. Our experiments utilized the AdamW optimizer with a cosine annealing learning rate schedule, setting the warm-up proportion to 0.1. The maximum learning rate for cosine annealing was set to 1e-4, while both the minimum and initial warm-up learning rates were set to 1e-6. We conducted LoRA fine-tuning and parallel training acceleration using the open-source LLM instruction fine-tuning library, llama-factory~\cite{zheng2024llamafactory}. All experiments were performed with a maximum of 8 training epochs and a batch size of 8, selecting the best-performing model on the validation set for testing. Detailed formats of the three instruction types are provided in the Appendix~\ref{app:instruct}.

\begin{table*}[t] %
\centering
\caption{Overall prediction performance of BehaviorLM compared with baselines. (a) Results on App Usage Dataset; (b) Results on Behavior Dataset.}
\label{tab:overall}

\resizebox{0.9\textwidth}{!}{%
\setlength{\tabcolsep}{5pt} %
\renewcommand{\arraystretch}{1.1} %
\begin{tabular}{clcccccc}
\toprule
\multicolumn{8}{c}{\textbf{(a) App Usage Dataset}} \\
\midrule
\textbf{Category} & \textbf{Model} & $Prec_w$ & $Rec_w$ & $Overall$ & $Head$ & $Medium$ & $Tail$ \\
\midrule
\multirow{2}{*}{Traditional}
& SASRec   & 0.5309 & 0.5759 & 0.2752 & 0.5255 & 0.2567 & 0.1733 \\
& Bert4Rec & 0.3452 & 0.5400 & 0.0962 & 0.5290 & -      & -      \\
\midrule
\multirow{3}{*}{LLM-Enhanced} 
& PITuning  & 0.5837 & 0.5133 & 0.2910 & 0.5029 & 0.3721 & 0.1066 \\
& LLM-ESR   & 0.5437 & 0.5906 & 0.2779 & 0.5615 & 0.2467 & 0.1750 \\
& AlphaFuse & 0.5621 & 0.6024 & 0.2873 & 0.5706 & 0.2552 & 0.1881 \\
\midrule
\multirow{6}{*}{LLM-Based}
& LLama-NT   & 0.5467 & 0.5346 & 0.3736 & 0.5335 & 0.3685 & 0.3000 \\
& GPT4o      & 0.5872 & 0.5678 & \underline{0.4557} & 0.5410 & \underline{0.4642} & \underline{0.4025} \\
& A-LLMRec   & 0.5908 & 0.6154 & 0.3514 & 0.5815 & 0.3539 & 0.2333 \\
& LLaRA      & 0.6074 & 0.6256 & 0.4455 & 0.5970 & 0.4468 & 0.3683 \\
& CoLLM      & 0.6053 & 0.6264 & 0.4478 & 0.6032 & 0.4423 & 0.3726 \\
& TALLRec    & \underline{0.6173} & \underline{0.6306} & 0.4397 & \underline{0.6205} & 0.4328 & 0.3580 \\
& \textbf{BehaviorLM} & \textbf{0.6440} & \textbf{0.6385} & \textbf{0.5514} & \textbf{0.6375} & \textbf{0.5308} & \textbf{0.5265} \\
\midrule
& Improv. & 4.3\% & 1.3\% & 21.0\% & 2.7\% & 14.3\% & 30.8\% \\
\bottomrule
\end{tabular}%
}

\vspace{0.2cm} %

\resizebox{0.9\textwidth}{!}{%
\setlength{\tabcolsep}{5pt}
\renewcommand{\arraystretch}{1.1}
\begin{tabular}{clcccccc}
\toprule
\multicolumn{8}{c}{\textbf{(b) Behavior Dataset}} \\
\midrule
\textbf{Category} & \textbf{Model} & $Prec_w$ & $Rec_w$ & $Overall$ & $Head$ & $Medium$ & $Tail$ \\
\midrule
\multirow{2}{*}{Traditional}
& SASRec   & 0.4818 & 0.5507 & 0.1531 & 0.4083 & 0.2159 & 0.0941 \\
& Bert4Rec & 0.1908 & 0.4368 & 0.0290 & 0.2500 & -      & -      \\
\midrule
\multirow{3}{*}{LLM-Enhanced} 
& PITuning  & 0.5829 & 0.5057 & 0.2121 & 0.5030 & 0.3545 & 0.0522 \\
& LLM-ESR   & 0.5325 & 0.5750 & 0.1633 & 0.4240 & 0.1772 & 0.0977 \\
& AlphaFuse & 0.5428 & 0.5857 & 0.1729 & 0.4335 & 0.1874 & 0.1093 \\
\midrule
\multirow{6}{*}{LLM-Based}
& LLama-NT   & 0.5091 & 0.4226 & 0.2433 & 0.4142 & 0.2467 & 0.2040 \\
& GPT4o      & 0.5735 & 0.5660 & \underline{0.3561} & 0.4575 & \underline{0.3703} & \underline{0.3248} \\
& A-LLMRec   & 0.5545 & 0.5856 & 0.2526 & 0.4567 & 0.3267 & 0.1601 \\
& LLaRA      & 0.5892 & 0.6099 & 0.3458 & 0.5175 & 0.3661 & 0.2932 \\
& CoLLM      & 0.5928 & 0.6125 & 0.3493 & \underline{0.5213} & 0.3674 & 0.2964 \\
& TALLRec    & \underline{0.5938} & \underline{0.6141} & 0.3425 & 0.5175 & 0.3636 & 0.2908 \\
& \textbf{BehaviorLM} & \textbf{0.6317} & \textbf{0.6402} & \textbf{0.4389} & \textbf{0.5558} & \textbf{0.4506} & \textbf{0.3978} \\
\midrule
& Improv. & 6.4\% & 4.3\% & 23.3\% & 6.6\% & 21.7\% & 22.5\% \\
\bottomrule
\end{tabular}%
}
\vspace{-0.3cm} 
\end{table*}

\subsection{Overall Performance}
We compare the performance of BehaviorLM with other baseline methods across two evaluation settings, and the results are summarized in Table~\ref{tab:overall}. Our key observations are as follows:
\begin{itemize}
    \item  \textbf{BehaviorLM consistently achieves the best performance on both datasets, as evaluated by various metrics.} Notably, under the \textit{long-tailed learning} evaluation protocol, BehaviorLM shows an improvement of up to {21.0\% }in Overall Accuracy for the App dataset and {23.3\% } for the Behavior dataset. This is primarily driven by a significant improvement in predicting tail behaviors, where BehaviorLM outperforms the best baseline by {30.8\% } (App) and {22.5\% } (Behavior). 
    
    \item \textbf{Fine-tuning LLMs for behavior prediction requires addressing the severe long-tailed distribution found in empirical data.} Existing LLM fine-tuning approaches for behavior prediction struggle with modeling this long-tailed distribution. LLM-based models such as LLaRA~\cite{liao2024llara} and TALLRec~\cite{bao2023tallrec}, which directly fine-tune the LLM on behavior data across all types, fail to outperform non-tuned GPT4o on the Medium and Tail categories. Similarly, LLM-enhanced methods that incorporate LLM-encoded knowledge into traditional models still do not achieve robust learning across both anchor and tail behaviors. In contrast, BehaviorLM employs a novel progressive tuning approach, enabling the LLM to first master predicting anchor behaviors and then generalize to the remaining tail behaviors.

    \item \textbf{The behavioral knowledge stored in LLMs proves highly beneficial for predicting user behaviors.} Traditional deep learning-based solutions, such as SASRec~\cite{kang2018self} and Bert4Rec~\cite{sun2019bert4rec}, struggle to compete with LLM-based prediction models because they cannot leverage this knowledge and perform poorly when fine-tuning data is limited. Additionally, Bert4Rec performs particularly poorly under long-tailed evaluation settings, failing to make accurate predictions on less frequently occurring behaviors (Medium and Tail categories).

\end{itemize}

\subsection{Investigating the Effect of Behavioral Knowledge}
Our evaluation in the previous subsection highlights the significant performance improvement brought by the LLM’s behavioral knowledge, as evidenced by the superior performance of LLM-based methods over traditional deep learning (DL) approaches. In this subsection, we conduct a more detailed analysis from this perspective. It is well-established that the more parameters an LLM contains, the greater its capacity to store knowledge learned from its pretraining corpus. Therefore, we examine the impact of model size on the LLM’s predictive capability by fine-tuning three versions of BehaviorLM with different backbones: Qwen-1.5B-v2, Llama-8B-v3.1, and Llama-70B-v3.1. {Here, the proportion of auxiliary task data is controlled at 5\% for all models.}

\begin{figure}[t]
\vspace{-1em}
\centering
\subcaptionbox{Anchor vs. Tail behavior}{
    \includegraphics[width=0.46\linewidth]{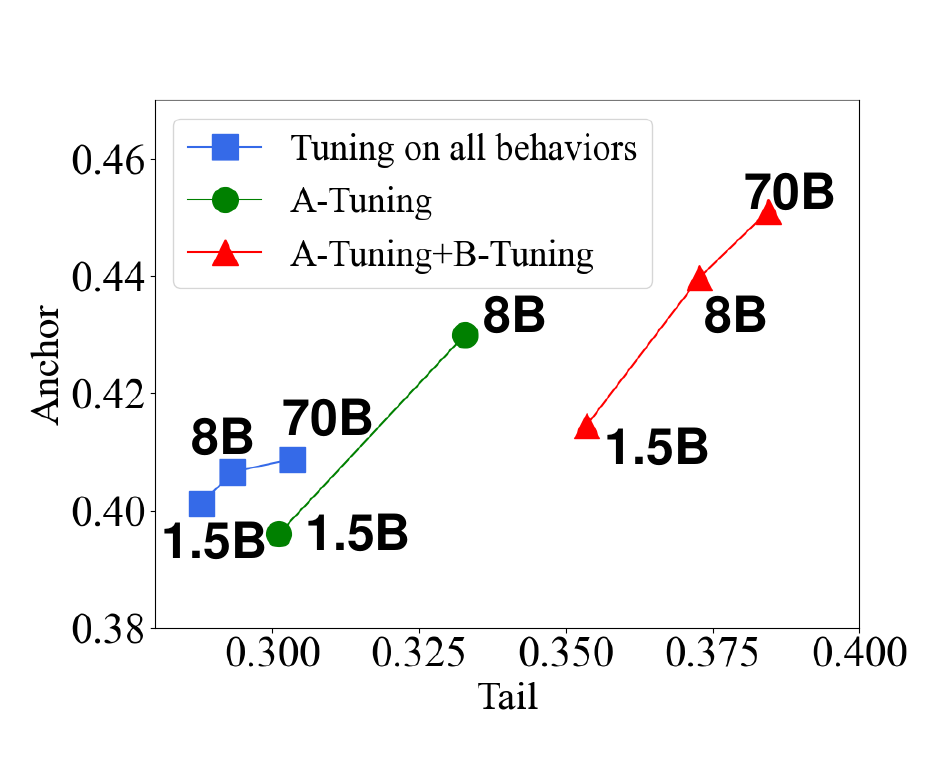}}
\subcaptionbox{Few-shot capability (8B)}{
    \includegraphics[width=0.46\linewidth]{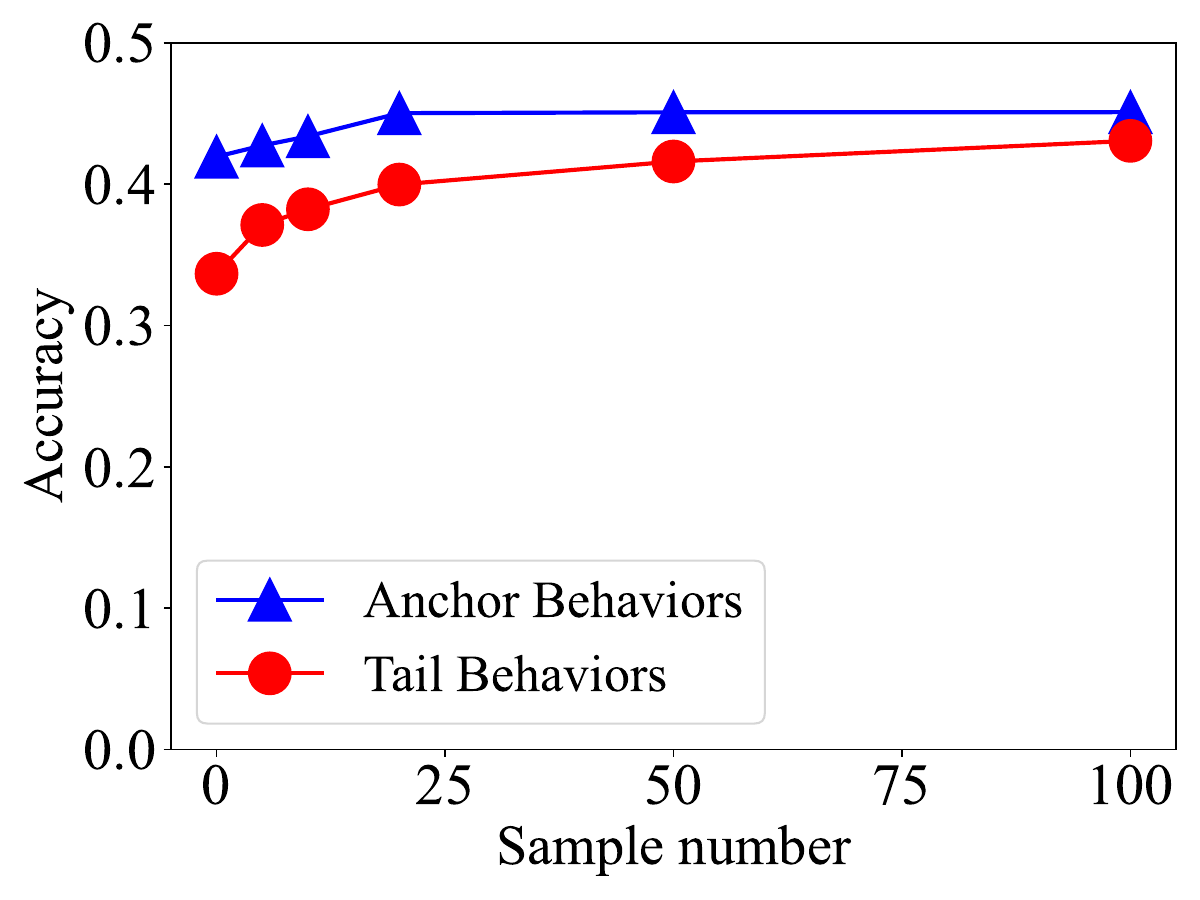 }}
\caption{The effect of behavioral knowledge under different model size (1.5B, 8B, 70B), in terms of performance robustness across behavior types and few-shot sample numbers. }
\label{fig:beh_knowledge_1}
\end{figure}

\subsubsection{Performance robustness.} We investigate whether this behavioral knowledge enables robust prediction across both anchor and tail behaviors, as well as under diverse few-shot settings.
\begin{itemize}
    \item In Figure~\ref{fig:beh_knowledge_1}(a), 
     we compare BehaviorLM's prediction performance on anchor and tail behaviors (Behavior dataset) using different model sizes and tuning strategies. Increasing the model size from 1.5B to 70B significantly improves behavior prediction accuracy. Specifically, the relative improvement is similar for anchor behaviors and tail behaviors, 
     demonstrating prediction robustness. Furthermore, when fine-tuning the LLM directly on all behaviors—without using our proposed progressive tuning approach (A-Tuning + B-Tuning)—the improvement from increasing model size is relatively marginal (blue curve in the figure). This suggests that our method effectively leverages the behavioral knowledge embedded in the LLM.
     
    \item In Figure~\ref{fig:beh_knowledge_1}(b), we vary the number of few-shot examples used in the B-tuning stage for BehaviorLM-8B and plot the performance curve. It is evident that with fewer than 20 examples per behavior type, BehaviorLM quickly learns to make robust predictions, demonstrating its ability to grasp behavior patterns efficiently even in low-data settings.
\end{itemize}

\subsubsection{Sample efficiency.} 
One significant advantage of leveraging the LLM's general behavioral knowledge is that it reduces the need for fine-tuning on large-scale user behavior data, demonstrating strong sample efficiency. To validate this, we compare BehaviorLM-1.5B and BehaviorLM-8B with another transformer-based model trained from scratch, using the objective function from SASRec~\cite{kang2018self} and the model architecture from GPT2~\cite{gpt2} (the \textit{small} version with 12 layers and 768 latent dimensions). Since the original Behavior dataset contains over 50 million events, we vary the sample size to observe its impact on prediction performance.
As shown in Figure~\ref{fig:beh_knowledge_2}, BehaviorLM demonstrates a significant improvement in sample efficiency. The transformer-based model trained from scratch only outperforms BehaviorLM when trained on nearly all 50 million samples, while BehaviorLM is fine-tuned on just 200,000 samples—over two orders of magnitude fewer. This highlights the remarkable sample efficiency advantage provided by the LLM's preexisting behavioral knowledge. Using 8xA100 (40G) GPUs, training BehaviorLM-8B on Behavior Dataset with 200k samples takes about 6 hours, while on APP Usage dataset (50000 samples) it takes no more than 2 hours. In contrast, training a SASRec model on a 50M sample dataset requires approximately 72 hours. Overall, although BehaviorLM has a lower training speed for each sample, it has a much higher sample efficiency than SASRec and thus takes much less time to achieve a good prediction accuracy.

\begin{figure}[t]

\centering
    \includegraphics[width=0.96\linewidth]{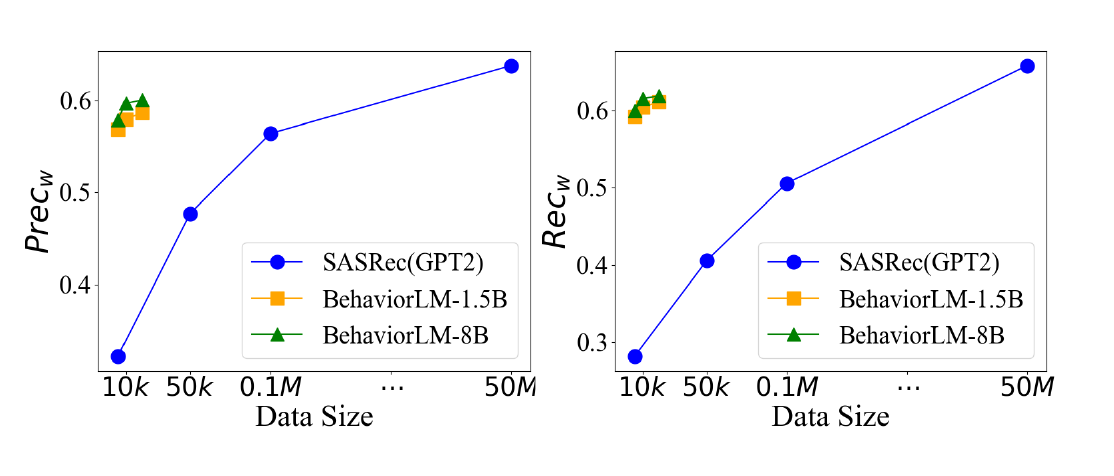 }
\caption{Comparison between BehaviorLM and a non-LLM transformer-based method under different sizes of training data.}
\label{fig:beh_knowledge_2}
\end{figure}

\subsection{Ablation Study}
In Table~\ref{tab:ablation_combined}, we evaluate the contribution of each design component to the overall performance through comprehensive ablation studies on both the App Usage and Behavior datasets. Specifically, we examine the performance drop when: 
(1) \textbf{w/o Aux. Task}: Removing the auxiliary conversation task during A-tuning, 
(2) \textbf{w/o DDS}: Replacing the difficulty-based data selection with uniform random selection during B-Tuning, 
(3) \textbf{w/o KMeans}: Removing the semantic clustering step (Difficulty-Weighted K-Means++), 
(4) \textbf{w/o DPO}: Replacing the Direct Preference Optimization in B-Tuning with standard Supervised Fine-Tuning (SFT), 
and (5) \textbf{w/o A-Tuning}: Skipping the first stage and fine-tuning the LLM directly on all behavior types.

\begin{table}[t]
    \centering
    \caption{Performance drop of ablations on App Usage and Behavior datasets.}
    \label{tab:ablation_combined}

    \resizebox{0.85\linewidth}{!}{
    \begin{tabular}{lcccc}
        \toprule
        \multicolumn{5}{c}{\textbf{(a) App Usage Dataset}} \\
        \midrule
        \textbf{BehaviorLM Variant} & \textit{Overall} & \textit{Head} & \textit{Medium} & \textit{Tail} \\
        \midrule
        A-Tuning w/o Aux. Task & -3.91\%  & -2.45\%  & -2.19\%  & -6.00\%  \\
        B-Tuning w/o DDS       & -7.90\%  & -5.74\%  & -6.30\%  & -10.11\% \\
        B-Tuning w/o Keans     & -2.76\%  & -1.60\%  & -2.95\%  & -3.52\%  \\
        B-Tuning w/o DPO       & -2.55\%  & -2.31\%  & -1.32\%  & -2.61\%  \\
        w/o A-Tuning           & -8.03\%  & -0.93\%  & -5.87\%  & -14.26\% \\
        \bottomrule
    \end{tabular}
    }

    \vspace{10pt} %

    \resizebox{0.85\linewidth}{!}{
    \begin{tabular}{lcccc}
        \toprule
        \multicolumn{5}{c}{\textbf{(b) Behavior Dataset}} \\
        \midrule
        \textbf{BehaviorLM Variant} & \textit{Overall} & \textit{Head} & \textit{Medium} & \textit{Tail} \\
        \midrule
        A-Tuning w/o Aux. Task & -4.57\%  & -1.10\%  & -6.11\%  & -4.40\%  \\
        B-Tuning w/o DDS       & -6.29\%  & -4.56\%  & -7.51\%  & -5.56\%  \\
        B-Tuning w/o Keans     & -3.08\%  & -0.90\%  & -4.43\%  & -4.65\%  \\
        B-Tuning w/o DPO       & -3.58\%  & -2.84\%  & -5.61\%  & -1.74\%  \\
        w/o A-Tuning           & -9.80\%  & -2.83\%  & -11.39\% & -10.93\% \\
        \bottomrule
    \end{tabular}
    }
\end{table}

The key conclusions from Table~\ref{tab:ablation_combined} are as follows:
First, all the design components contribute to the final prediction performance. For head-category behaviors, the difficulty-based data selection strategy in B-tuning has the most significant impact, as it improves the model's ability to differentiate between different behaviors, avoiding the erroneous and significant suppression of head-category behavior predictions. For medium and tail categories, the largest contribution comes from the A-tuning stage, without which prediction accuracy drops significantly. This is expected, as fine-tuning the LLM directly on all behaviors causes a loss in predictive capability for tail behaviors, which cannot be recovered through few-shot tuning in the B-tuning stage. Furthermore, \textit{w/o KMeans} shows a moderate decline, validating that while difficulty is crucial, ensuring semantic diversity through clustering prevents the model from overfitting to specific types of failure cases. Replacing DPO with SFT (\textit{w/o DPO}) results in a consistent performance drop, suggesting that the contrastive nature of DPO better leverages the hard negatives generated during the difficulty scoring phase. Finally, removing the auxiliary task (\textit{w/o Aux. Task}) causes a moderate decline, proving that maintaining general capabilities via multi-task learning acts as a necessary regularizer to support robust behavior prediction.

\begin{figure}[t]
\centering
    \includegraphics[width=0.5\linewidth]{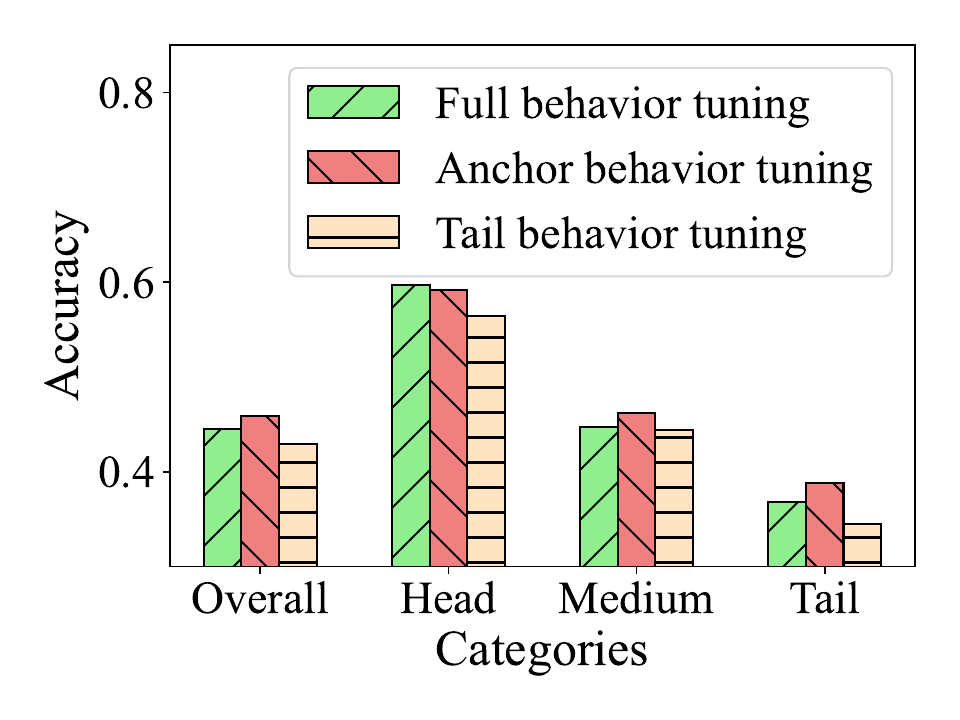 }
\hspace{-1em}
    \includegraphics[width=0.5\linewidth]{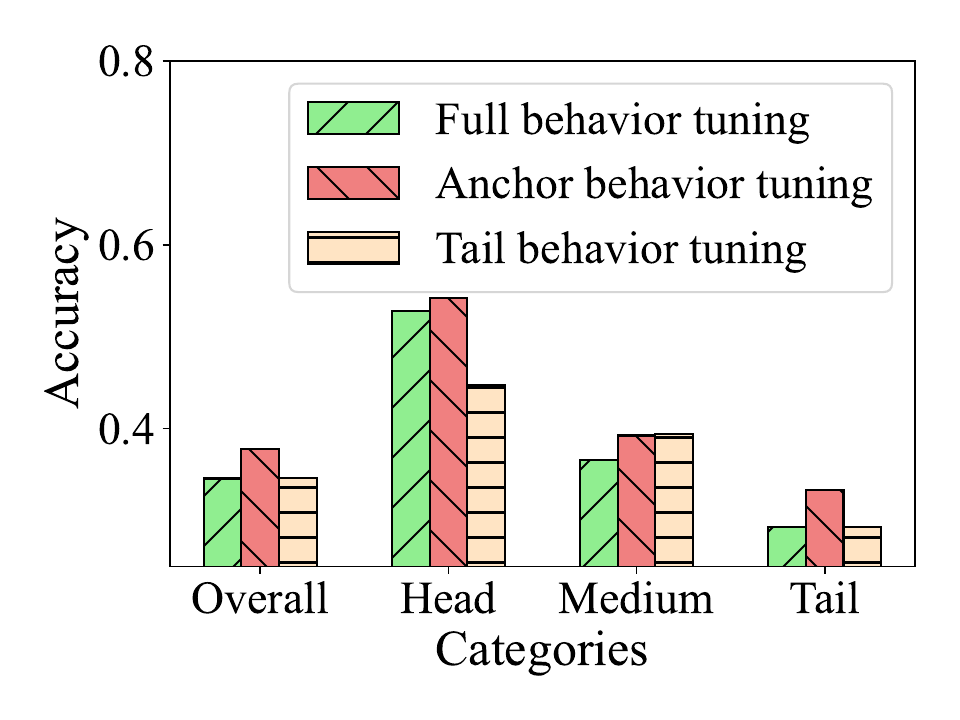 }
    \vspace{-3mm}
\caption{Performance comparison between fine-tuning on all behaviors, anchor behaviors and tail behaviors. }
\label{fig:anchor}
\end{figure}

\subsection{Investigating the Necessity of Anchor Behavior Tuning}

We have demonstrated the importance of progressive tuning for LLM-based behavior prediction. However, the necessity of first tuning on anchor behaviors remains unclear.
To address this, we replace the A-tuning stage with a tuning stage focused solely on tail behaviors(here, we also include medium-frequency behaviors, as tail behaviors are relatively rare).
As illustrated in Figure~\ref{fig:anchor}, this alternative approach leads to poorer prediction performance in both anchor and tail behaviors. 
This suggests that, since anchor behaviors represent the core structure of human daily life, prompting the LLM to follow a curriculum from anchor behaviors to tail behaviors is more effective than the reverse.

Additionally, we conducted hyperparameter experiments on the Behavior dataset to assess the impact of varying auxiliary task data proportions. As evident from Table~\ref{Hyperparameter experiment}, an appropriate amount of auxiliary task data can significantly enhance the model's performance. Insufficient auxiliary task data fails to deliver notable improvements, whereas an excessive amount can disrupt the model's predictive capabilities.

\begin{table}[h]
  \centering
  \caption{The Hyperparameter Experiment of Auxiliary Task Data Ratio on  Behavior Dataset }
  \label{Hyperparameter experiment}
  \scalebox{1.2}{ %
  \begin{tabular}{ccccc}
    \toprule
    Auxiliary Task Data Ratio          & overall      &   head   &medium & tail             \\
    \midrule
    0\%   & 0.404 & 0.556 & 0.418 & 0.376 \\
    2.5\% & 0.423 & 0.543 & 0.432 & 0.386 \\
    5\%   & 0.439 & 0.556 & 0.452 & 0.398 \\
    10\%  & 0.416 & 0.542 & 0.423 & 0.382 \\
    20\%  & 0.410 & 0.522 & 0.414 & 0.348 \\
    \bottomrule
  \end{tabular}
  \label{tab:three-line}
  }
\end{table}

\section{Extended Hyperparameter and Mechanism Analysis}
\label{app:extended_analysis}
In this section, we provide a deeper analysis regarding the hyperparameters used for data partition and investigate the underlying mechanism of the auxiliary task utilized in the A-Tuning stage.

\subsection{Sensitivity to Anchor/Tail Threshold}
\label{app:threshold_sensitivity}
In our main experiments, we defined ``Anchor Behaviors'' as those occurring with a frequency greater than 1\%. To evaluate the robustness of BehaviorLM with respect to this threshold selection, we conducted additional experiments on the \textit{Behavior Dataset}. We varied the threshold for distinguishing anchor and tail behaviors to 0.5\%, 1\% (the setting used in the main paper), and 2\%.

The results are presented in Table~\ref{tab:threshold_sensitivity}. As observed, BehaviorLM achieves consistent performance across different threshold settings. For instance, the Overall Accuracy remains stable around 0.42, and the Tail Accuracy shows only minor fluctuations. This demonstrates that our method is robust to the specific choice of the long-tail threshold and does not rely on a specific data partition to achieve superior performance.

\begin{table}[h]
    \centering
    \caption{Performance comparison under different Anchor/Tail division thresholds on the Behavior Dataset.}
    \label{tab:threshold_sensitivity}
    \resizebox{0.8\textwidth}{!}{
    \begin{tabular}{lcccccc}
        \toprule
        Threshold &  $Prec_w$ &$Rec_w$ &$Overall$ &$Head$ &$Medium$ &$Tail$ \\
        \midrule
    0.5\%          & 0.6301 & 0.6429 & 0.4449 & 0.5394 & 0.4382 & 0.4202 \\
    1.0\% (Default) & 0.6317 & 0.6402 & 0.4389 & 0.5558 & 0.4506 & 0.3978 \\
    2.0\%          & 0.6250 & 0.6368 & 0.4322 & 0.5445 & 0.4304 & 0.3973 \\
        \bottomrule
    \end{tabular}
    }
\end{table}

\subsection{Investigation into the Mechanism of Auxiliary Tasks}
\label{app:auxiliary_mechanism}

To improve the model's generalization capability during the A-Tuning stage, we incorporated a general conversational dataset (ShareGPT) as an auxiliary task. A key question arises regarding the mechanism of this improvement: does it stem from specific knowledge transfer or a general regularization effect?
We hypothesize that the auxiliary dialogue task functions primarily through a \textbf{regularization effect}. By maintaining the LLM's general text generation capabilities, it prevents overfitting to the anchor behavior prediction task and mitigates the catastrophic forgetting of common sense.

To validate this hypothesis, we performed experiments on the \textit{Behavior Dataset} replacing the original ShareGPT data with two alternative domains:
\begin{itemize}
    \item \textbf{MATH} \cite{hendrycks2021measuring}: A dataset focused on mathematical reasoning problems.
    \item \textbf{HealthQA} \cite{hosseini2024benchmark}: A medical dialogue dataset that is semantically unrelated to the user behavior domain.
\end{itemize}
The results, summarized in Table~\ref{tab:aux_task_type}, show that model performance does not differ significantly across the three diverse auxiliary tasks. Whether using general chat, mathematical reasoning, or medical dialogue, the improvement in tail behavior prediction remains comparable. This reinforces our conclusion that the benefit of the auxiliary task stems from general regularization---helping the model maintain its intrinsic reasoning and generation abilities---rather than domain-specific semantic knowledge transfer.

\begin{table}[h]
    \centering
    \caption{Performance comparison using different types of auxiliary datasets on the Behavior Dataset.}
    \label{tab:aux_task_type}
    \resizebox{0.8\textwidth}{!}{
    \begin{tabular}{lcccccc}
        \toprule
        Auxiliary Task Type & $Prec_w$ &$Rec_w$ &$Overall$ &$Head$ &$Medium$ &$Tail$ \\
        \midrule
    ShareGPT (Default) & 0.6317 & 0.6402 & 0.4389 & 0.5558 & 0.4506 & 0.3978 \\
    MATH               & 0.6293 & 0.6344 & 0.4451 & 0.5471 & 0.4328 & 0.4224 \\
    HealthQA           & 0.6330 & 0.6355 & 0.4517 & 0.5481 & 0.4448 & 0.4267 \\
        \bottomrule
    \end{tabular}
    }
\end{table}

\subsection{Generalization and Robustness}
\label{sec:generalization_robustness}

To assess the generalization capability of BehaviorLM on long-tailed tasks beyond smartphone usage, we extended our evaluation to the \textbf{Foursquare-NYC} dataset \cite{yang2014modeling}, a representative benchmark in the mobility prediction domain. Furthermore, we examined the robustness of our method with respect to hyperparameter settings. It is worth noting that we employed the \textbf{identical configuration} for the Foursquare-NYC dataset as used in the main experiments on App Usage and Behavior datasets, without performing separate hyperparameter optimization. Specifically, we fixed the anchor threshold at 1\%, the auxiliary task ratio at 5\%, and the number of samples per behavior type for B-Tuning at 20.

As presented in Table~\ref{tab:foursquare_res}, BehaviorLM consistently outperforms all baselines across all metrics. Notably, in the tail segment, our method achieves a Tail Accuracy of \textbf{0.2157}, significantly surpassing the best-performing baseline (GPT4o with 0.1405), confirming that our progressive tuning strategy effectively generalizes to other long-tailed domains. And the consistent performance achieved under these fixed settings across three diverse datasets demonstrates that BehaviorLM is highly robust and does not rely on extensive, dataset-specific hyperparameter tuning.

\begin{table}[h]
    \centering
    \caption{Overall prediction performance on the Foursquare-NYC dataset.}
    \label{tab:foursquare_res}
    \resizebox{0.8\columnwidth}{!}{
    \begin{tabular}{lcccccc}
        \toprule
        \textbf{Model} & \textbf{$Prec_w$} & \textbf{$Rec_w$} & \textbf{Overall} & \textbf{Head} & \textbf{Medium} & \textbf{Tail} \\
        \midrule
        SASRec & 0.2781 & 0.2841 & 0.1307 & 0.2202 & 0.0980 & 0.0065 \\
        Bert4Rec & 0.2080 & 0.2445 & 0.0556 & 0.1205 & 0.0000 & 0.0000 \\
        PITuning & 0.3243 & 0.3167 & 0.1220 & 0.2132 & 0.1225 & 0.0000 \\
        LLM-ESR & 0.2834 & 0.2896 & 0.1430 & 0.2238 & 0.1085 & 0.0152 \\
        AlphaFuse & 0.3313 & 0.3217 & 0.1463 & 0.2246 & 0.1493 & 0.0240 \\
        LLama-NT & 0.2822 & 0.2826 & 0.1514 & 0.1863 & 0.1372 & 0.1144 \\
        GPT4o & 0.2989 & 0.3010 & 0.1912 & 0.2273 & 0.2056 & 0.1405 \\
        A-LLMRec & 0.2819 & 0.2769 & 0.0880 & 0.1520 & 0.0932 & 0.0882 \\
        LLaRA & 0.3248 & 0.3245 & 0.1786 & 0.2206 & 0.2059 & 0.1046 \\
        CoLLM &0.3287 & 0.3196 & 0.1824 & 0.2157 & 0.2069 & 0.1075\\
        TALLRec & 0.3246 & 0.3220 & 0.1786 & 0.2181 & 0.2157 & 0.1013 \\
        \midrule
        \textbf{BehaviorLM} & \textbf{0.3508} & \textbf{0.3471} & \textbf{0.2386} & \textbf{0.2525} & \textbf{0.2451} & \textbf{0.2157} \\
        \bottomrule
    \end{tabular}
    }
\end{table}

\section{Related Works}

\subsection{LLM-Enhanced Behavior Prediction Models}

User behavior prediction based on the most recent $L$ events is similar to sequential recommendation. The key difference is that behavior prediction focuses on recurring daily actions, while item recommendation emphasizes novel content. Due to limited research in behavior prediction, we draw on related work in sequential recommendation.
Recent studies in recommendation explore knowledge alignment between language and recommendation domains. For example, SAID\cite{hu2024enhancing} proposed improving sequential recommendation via LLM-based semantic embedding learning, PLM-Rec applied mutual information maximization\cite{geng2022path} , PITuning introduced PITuning for cross-modal pattern extraction\cite{gong2024population}, and LLM-ESR proposed a cross-attention mechanism for sequence alignment\cite{liu2024llm}. In addition, to address the issue of sparse user data, ProEx\cite{zhang2025proex} utilizes LLMs for profile extrapolation to generate enriched user representations, augmenting the input for downstream recommendation models. LLMCDSR\cite{xin2025llmcdsr} leverages LLMs to transfer knowledge and enhance user interest modeling in sparse domains, effectively mitigating the cold-start problem for traditional predictors.
While these methods enhance traditional models by aligning language and recommendation knowledge, they underutilize LLMs' zero-shot and few-shot generalization. Our approach addresses this gap with progressive fine-tuning, preserving general behavioral knowledge while improving prediction of infrequent long-tail behaviors without sacrificing performance on frequent ones.

\subsection{LLM-Based Behavior Prediction Models}
Current behavior prediction models largely rely on embedding neural networks, which often function as black-box models with limited interpretability. LLMs, with their vast world knowledge and powerful reasoning abilities, offer an alternative approach that enhances interpretability in user behavior prediction~\cite{wu2024survey}. The introduction of LLMs into behavior prediction was pioneered by \citet{bao2023tallrec}, who demonstrated the impressive few-shot performance of LLMs in the recommendation domain. To formalize this paradigm, \citet{zhang2025recommendation} proposed "Recommendation as Instruction Following," establishing a unified framework where the LLM directly outputs items based on natural language instructions.
Building on these foundations, several strategies have been proposed to optimize LLM performance. \citet{liao2024llara} employed a curriculum learning strategy to progressively fine-tune LLMs from easier tasks to more complex ones, while \citet{kimlarge} utilized multi-task training for embedding alignment. Furthermore, \citet{mao2025reinforced} applied reinforcement learning to dynamically optimize prompts, enhancing the LLM's ability to capture personalized user intent.
Regarding complex data distributions, \citet{wu2024coral} introduced CoRAL, a retrieval-augmented LLM framework that explicitly improves long-tail recommendation by retrieving collaborative evidence. Similarly, \citet{xia2025hierarchical} proposed a hierarchical tree search mechanism based on LLMs to model lifelong user behaviors.
Other approaches focus on specific modeling aspects. For instance, the Chain-of-Planned Behavior framework by \citet{shao2024beyond} captures the spatial-temporal dynamics of user activities. \citet{lin2024bridging} introduced a Transition paradigm combining multiple identifiers to enhance recommendations, and \citet{lei2024recexplainer} proposed alignment techniques to train LLMs to mimic traditional recommender models for customizable explanations.

However, these methods primarily focus on converting behavior sequences into textual representations for LLM training or enhancing external information retrieval, often neglecting the critical differences between anchor and tail behaviors. This limits their zero-shot generalization ability. Our proposed approach, BehaviorLM, differentiates them by addressing the long-tailed distribution of behaviors. 
First, we fine-tune on anchor behaviors while preserving general behavioral knowledge, and second, we fine-tune on a balanced subset of behaviors based on sample difficulty, significantly enhancing the prediction of tail behaviors without sacrificing anchor behavior performance.

\section{Conclusion}

In this paper, we leverage the rich behavioral knowledge in LLMs to tackle user behavior prediction, with a focus on long-tail behavior prediction. We propose a progressive tuning approach, where the LLM first learns frequent anchor behaviors before generalizing to rarer tail behaviors. Experiments on two real-world datasets show that BehaviorLM outperforms state-of-the-art methods, achieving up to {30.8\%/22.5\%} improvement in long-tail behavior prediction, addressing a traditionally challenging aspect of behavior modeling. Analysis highlights that addressing the long-tailed behavior distribution is essential for effectively utilizing LLMs' behavioral knowledge in fine-tuning.

\appendix
\section{Details of Used Datasets}
\label{app:dataset}

Detailed statistics of the Behavior Dataset are shown in the figure~\ref{app:dataset_fig}.

\begin{figure}[!htbp]
    \centering
    
    \begin{subfigure}[b]{0.4\textwidth}
        \centering
        \includegraphics[width=\linewidth]{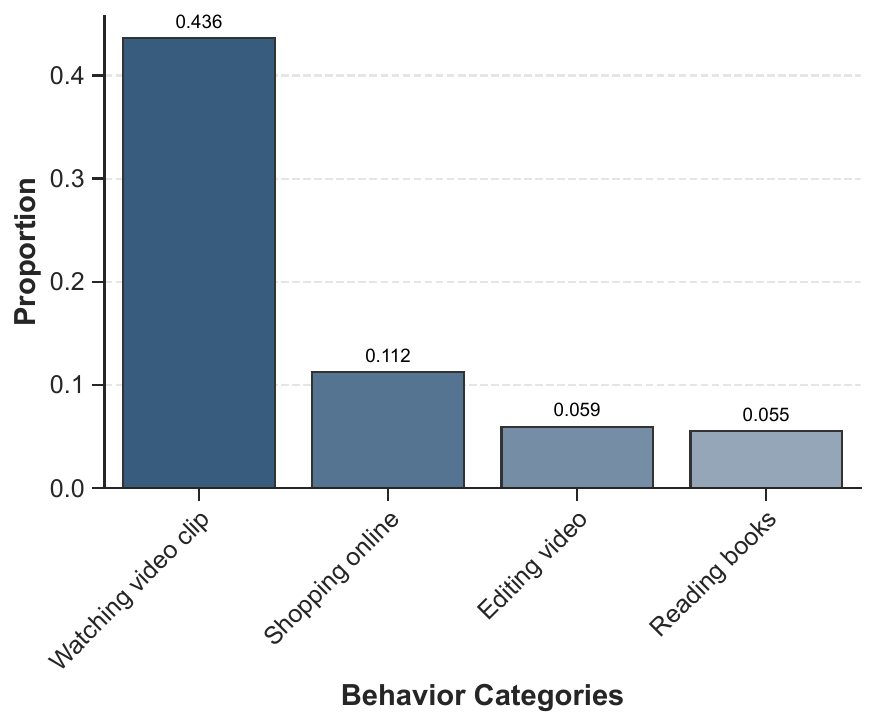}
        \caption{High frequency behaviors}
        \label{fig:high_freq}
    \end{subfigure}

    \begin{subfigure}[b]{0.65\textwidth}
        \centering
        \includegraphics[width=\linewidth]{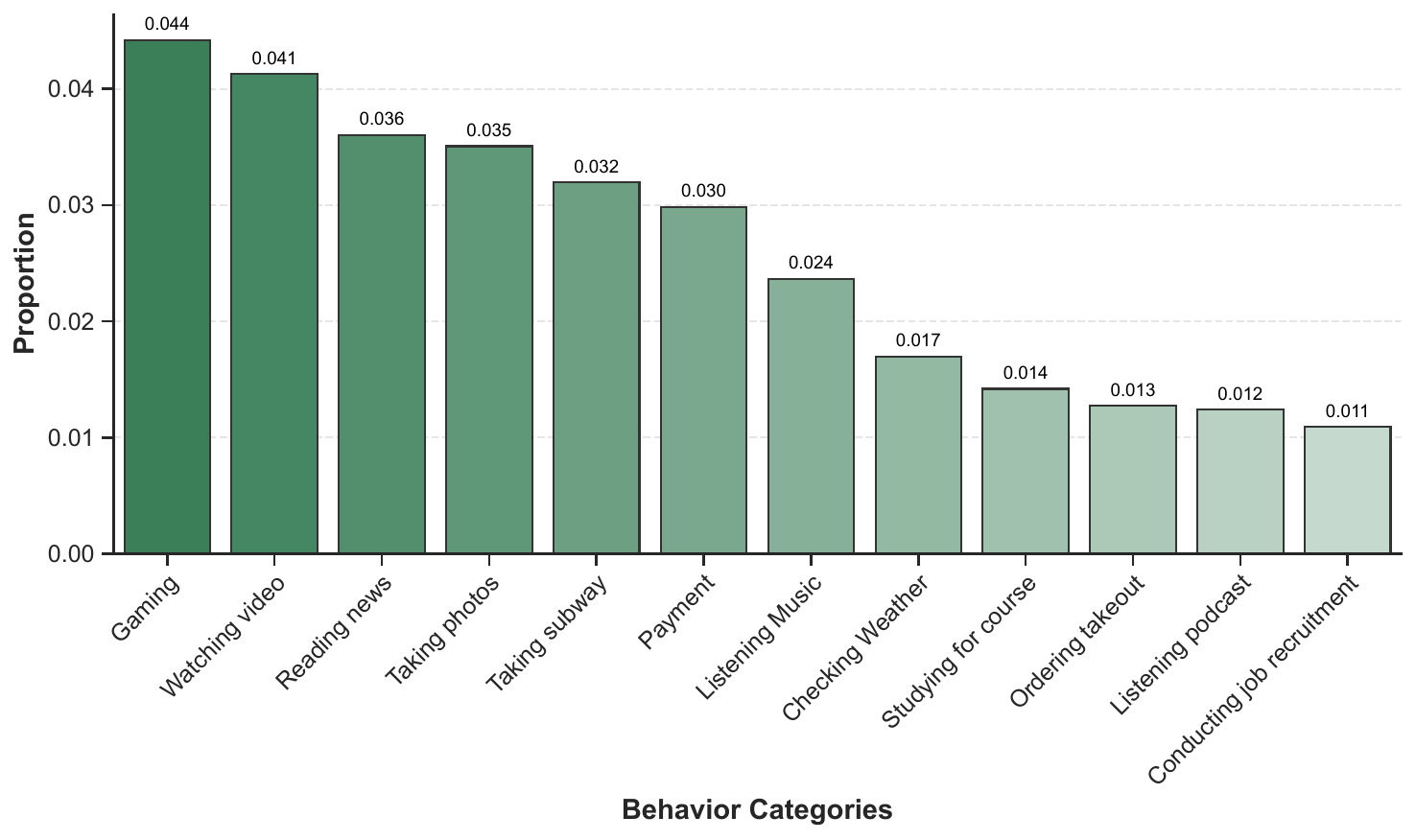}
        \caption{Medium frequency behaviors}
        \label{fig:mid_freq}
    \end{subfigure}

    \begin{subfigure}[b]{0.8\textwidth}
        \centering
        \includegraphics[width=\linewidth]{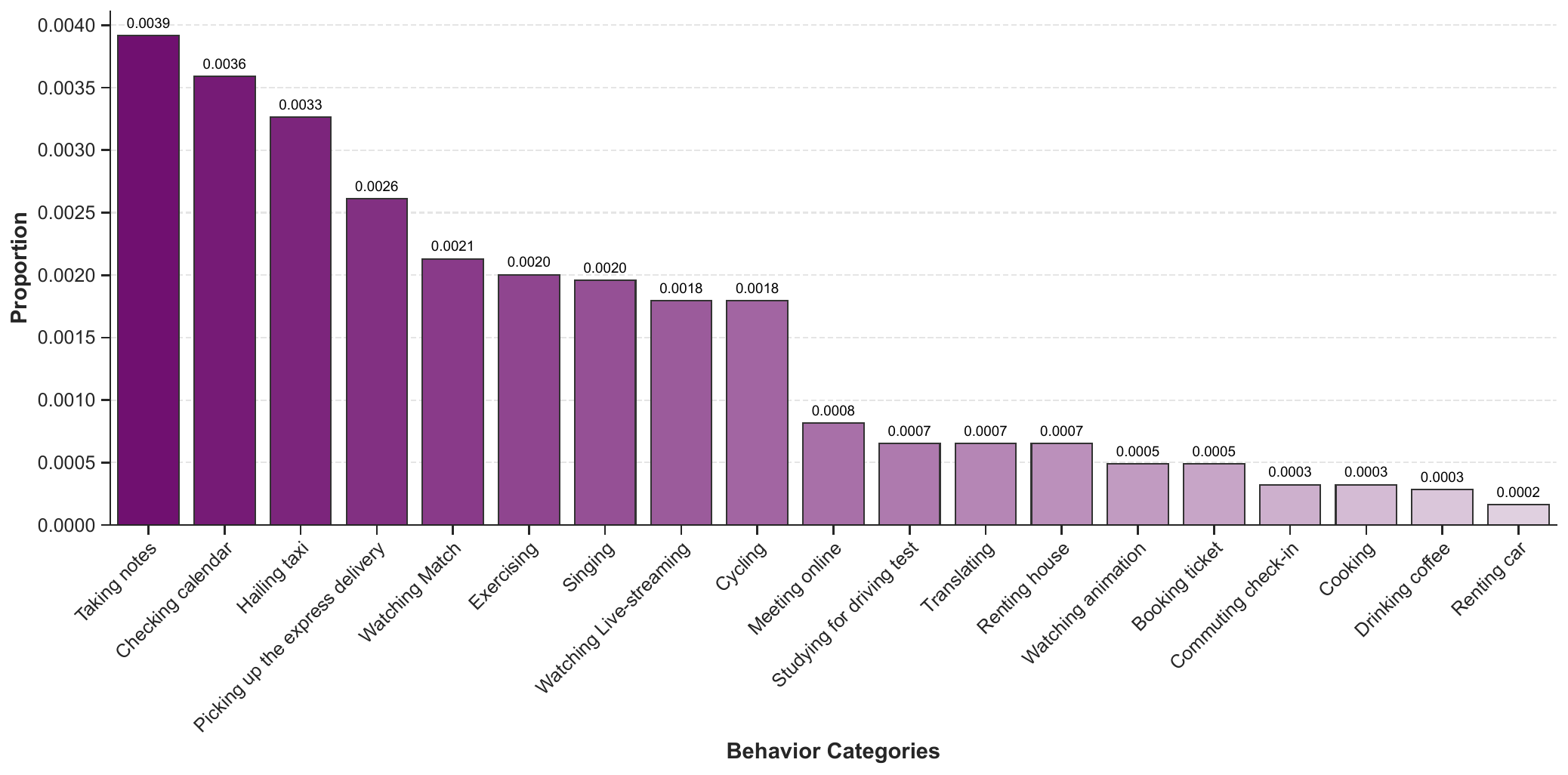}
        \caption{Low frequency behaviors}
        \label{fig:low_freq}
    \end{subfigure}
    
    \caption{Detailed statistics of the Behavior Dataset. The behaviors are categorized into (a) high, (b) medium, and (c) low frequencies based on their occurrence proportions.}
    \label{app:dataset_fig}
\end{figure}

\section{Details of Used Metrics}
\label{app:metric}
The formula for $Prec_w$ :
\begin{equation}\label{equ:Prec_w}
    Prec_w = \frac{\sum_{c \in C} (\text{TP}_c + \text{FP}_c) \cdot \text{Precision}_c}{\sum_{c \in C} (\text{TP}_c + \text{FP}_c)}
\end{equation}
The formula for $Rec_w$ :
\begin{equation}\label{equ:Rec_w}
    Rec_w = \frac{\sum_{c \in C} (\text{TP}_c + \text{FN}_c) \cdot \text{Recall}_c}{\sum_{c \in C} (\text{TP}_c + \text{FN}_c)}
\end{equation}
The formula for $Overall$ :
\begin{equation}\label{equ:Rec_m}
    Accuracy = \frac{1}{|C|} \sum_{c \in C} \frac{\text{TP}_c}{\text{TP}_c + \text{FN}_c}
\end{equation}
The formula for $Head$ :
\begin{equation}\label{equ:Rec_m}
    Accuary = \frac{1}{|C_h|} \sum_{c \in C_h} \frac{\text{TP}_c}{\text{TP}_c + \text{FN}_c}
\end{equation}
The formula for $Medium$ :
\begin{equation}\label{equ:Rec_m}
    Accuary = \frac{1}{|C_m|} \sum_{c \in C_m} \frac{\text{TP}_c}{\text{TP}_c + \text{FN}_c}
\end{equation}
The formula for $Tail$ :
\begin{equation}\label{equ:Rec_m}
    Accuary = \frac{1}{|C_t|} \sum_{c \in C_t} \frac{\text{TP}_c}{\text{TP}_c + \text{FN}_c}
\end{equation}

Where $|C|$ represents the total number of classes, $|C_h|$ represents the total number of classes belonging to the head category,
Where $|C_m|$ represents the total number of classes belonging to the medium category,
Where $|C_h|$ represents the total number of classes belonging to the tail category.
True Positives $\text({TP}_c)$ denotes the number of samples correctly classified as class $c$, False Positives $\text({FP}_c)$ represents the number of samples incorrectly classified as class $c$, and False Negatives $\text({FN}_c)$ stands for the number of samples incorrectly classified as other classes instead of class $c$. And $\text{Precision}_c$ and $\text{Recall}_c$ respectively refer to the precision and recall of class $c$.\\

\vspace{-2em}

\section{Baselines}
\label{app:baselines}

\textbf{SASRec~\cite{kang2018self}.}
uses self-attention mechanisms to model user behavior sequences. It captures both short-term and long-term dependencies in sequential data, allowing it to focus on the most relevant items in the user's interaction history for recommendation.

\textbf{BERT4Rec~\cite{sun2019bert4rec}.}
models user behavior sequences using deep bidirectional self-attention. By jointly considering the context before and after an item, it predicts the randomly masked items within the sequence, achieving excellent predictive performance.

\textbf{CoLLM~\citep{zhang2025collm}}
captures collaboration information using external traditional models and maps it into the LLM's input embedding space as collaboration embeddings. This external integration allows effective modeling of collaboration without modifying the LLM, enabling flexible use of various collaboration modeling techniques.

\textbf{LLaRa~\cite{liao2024llara}}
introduces a hybrid prompting method that integrates both world knowledge and behavioral patterns into item representations. It conducts curriculum prompt tuning to achieve modality alignment.

\textbf{A-LLMRec~\cite{kimlarge}}
bridges the knowledge between the language and recommendation domains by training an alignment network with a variety of tasks, targeting both warm and cold-start scenarios.

\textbf{PITuning~\cite{gong2024population}}
loads pre-trained Large Language Model (LLM) parameters to acquire textual knowledge and then designs an adaptive unlearning strategy to address the long-tail preference issue, achieving excellent performance in user behavior prediction.

\textbf{LLM-ESR~\cite{liu2024llm}}
{leverages the semantic information from LLMs, proposes a dual-view modeling framework enhanced through embedding techniques to capture the nuances of long-tail items better, demonstrating strong performance across multiple datasets.}.

\textbf{TALLRec~\cite{bao2023tallrec}}
is one of the earlier methods to integrate Large Language Models (LLMs) with the recommendation domain. It employs a two-stage tuning process—Alpaca Tuning and Rec-Tuning—to finetune LLMs for recommendations, enabling effective and efficient adaptation of LLMs with only a small number of tuning samples.

\textbf{AlphaFuse~\cite{hu2025alphafuse}}
is a simple yet effective language-guided learning strategy that addresses long-tail intent modeling by learning ID embeddings within the null space of language embeddings.

For comparison, we also consider two LLMs that are not fine-tuned on behavioral data, i.e., GPT4o~\cite{achiam2023gpt} and LLama3.1-8B~\cite{dubey2024llama}.

\section{Details of Used Instructions}
\label{app:instruct}

\textbf{Instruct1:} This user has done behaviors [HistoryHere] in the previous. Day of the week, the hour, and the place of the next behavior are [next-intent-info], respectively. Choose the answer from the following behavior candidate set: [CansHere]. The answer is [Output]. \\
\textbf{Instruct2:} The user's historical behavior information sequence is: [HistoryHere]. Day of the week, the hour, and the place of the next behavior are [next-intent-info], respectively. Given the following behavior candidate set: [CansHere], recommend one intention for this user to do next. The intent you recommend is [Output].  \\
\textbf{Instruct3:} The behavior history of this user is: [HistoryHere].Day of the week, the hour, and the place of the next behavior are [next-intent-info, respectively. Recommend a next intention for this user to do from the following behavior candidate set: [CansHere].The recommendation is [Output].

\bibliographystyle{ACM-Reference-Format}
\balance
\bibliography{reference}

\appendix

\end{document}